\documentclass{article}

\usepackage{xcolor}         
\usepackage{graphicx,multirow}
\usepackage{amsmath}
\usepackage{enumitem}
\usepackage[linesnumbered,ruled,vlined, noend]{algorithm2e}
\usepackage{amsfonts}
\usepackage{tikz}

\usepackage{microtype}
\usepackage{graphicx}
\usepackage{subfigure}
\usepackage{booktabs} 

\usepackage[pagebackref,breaklinks,colorlinks,allcolors=cvprblue]{hyperref}

\usepackage{caption}

\usepackage[accepted]{icml2025}

\usepackage{amsmath}
\usepackage{amssymb}
\usepackage{mathtools}
\usepackage{amsthm}

\usepackage[capitalize,noabbrev]{cleveref}

\theoremstyle{plain}

\theoremstyle{definition}

\theoremstyle{remark}

\usepackage[textsize=tiny]{todonotes}

\definecolor{YellowGreen}{rgb}{0.60, 0.80, 0.20}

\icmltitlerunning{}

\begin{document}

\twocolumn[
\icmltitle{TV-Dialogue: Crafting Theme-Aware Video Dialogues with\\ Immersive Interaction}



\icmlsetsymbol{equal}{*}

\begin{icmlauthorlist}
\icmlauthor{Sai Wang}{whu}
\icmlauthor{Fan Ma}{zju}
\icmlauthor{Xinyi Li}{whu}
\icmlauthor{Hehe Fan}{zju}
\icmlauthor{Yu Wu}{whu}
\end{icmlauthorlist}

\centering{\href{https://wangsai23.github.io/TV-Dialogue/}{https://wangsai23.github.io/TV-Dialogue/}}

\icmlaffiliation{whu}{School of Computer Science, Wuhan University}
\icmlaffiliation{zju}{Zhejiang University}

\icmlcorrespondingauthor{Yu Wu}{wuyucs@whu.edu.cn}

\icmlkeywords{Machine Learning, ICML}

\vskip 0.3in
]



\printAffiliationsAndNotice{}  

\begin{abstract}
Recent advancements in LLMs have accelerated the development of dialogue generation across text and images, yet video-based dialogue generation remains underexplored and presents unique challenges. 
In this paper, we introduce Theme-aware Video Dialogue Crafting (TVDC), a novel task aimed at generating new dialogues that align with video content and adhere to user-specified themes. 
We propose TV-Dialogue, a novel multi-modal agent framework that ensures both theme alignment (i.e., the dialogue revolves around the theme) and visual consistency (i.e., the dialogue matches the emotions and behaviors of characters in the video) by enabling real-time immersive interactions among video characters, thereby accurately understanding the video content and generating new dialogue that aligns with the given themes.
To assess the generated dialogues, we present a multi-granularity evaluation benchmark with high accuracy, interpretability and reliability, demonstrating the effectiveness of TV-Dialogue on self-collected dataset over directly using existing LLMs.
Extensive experiments reveal that TV-Dialogue can generate dialogues for videos of any length and any theme in a zero-shot manner without training.
Our findings underscore the potential of TV-Dialogue for various applications, such as video re-creation, film dubbing and its use in downstream multimodal tasks.
\end{abstract}

\begin{figure}[t]
\centering    
\includegraphics[width=0.45\textwidth]{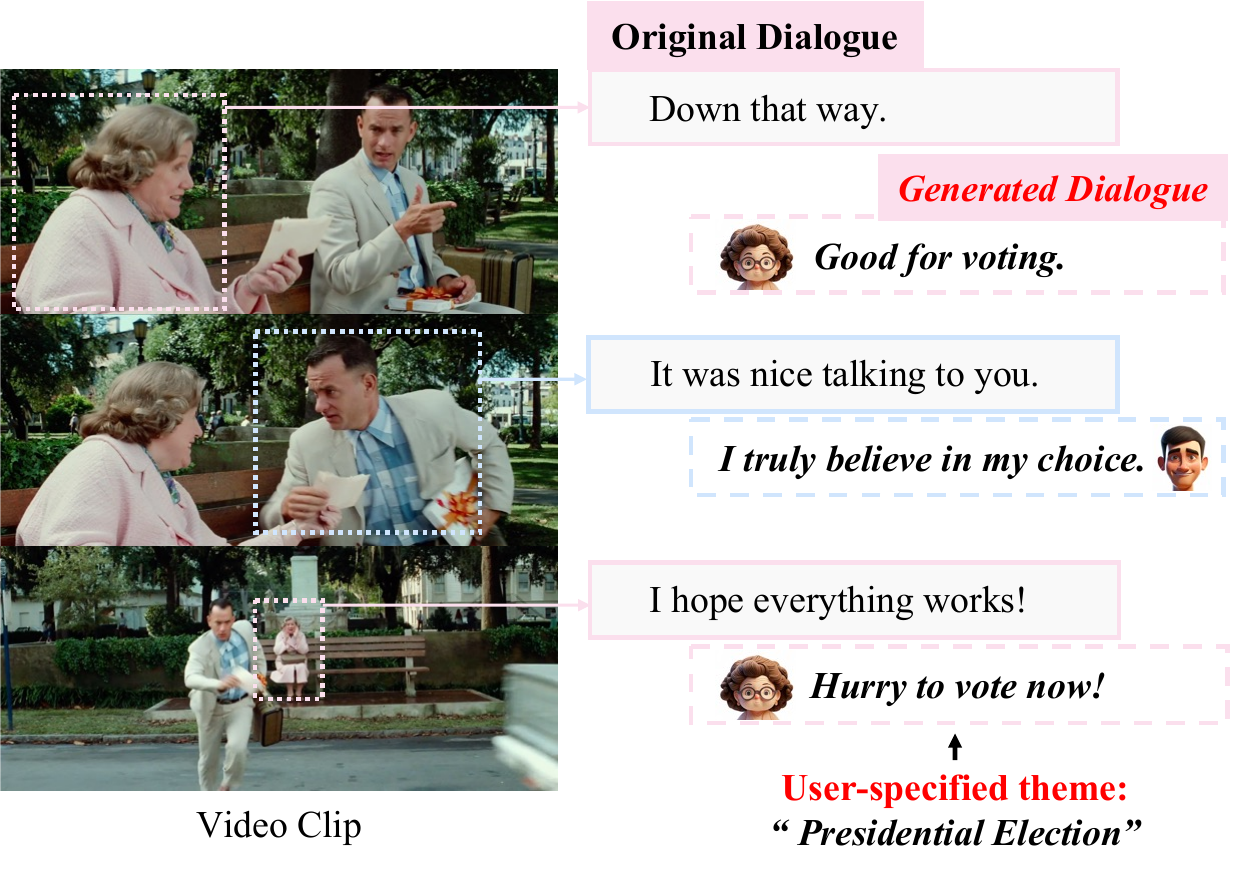}
\vspace{-0.2in}
\caption{
Given an arbitrary user-specified theme, the \textbf{Theme-aware Video Dialogue Crafting (TVDC)} task seeks to generate novel dialogues aligned with video content and theme.
The solid box represents the original dialogue, while the dashed box represents the new dialogue about the ``presidential election''.
}    
\vspace{-0.2in}
\label{fig:intro}
\end{figure}

\section{Introduction}

Recent advances in large language models (LLMs)~\cite{zhao2023survey, chang2024survey} have significantly boosted the development of dialogue-generated content, e.g. text-based dialogue generation~\cite{yarats2018hierarchical,li-etal-2016-deep,huang2018automatic} and image-based dialogue generation~\cite{yang2021open,sun-etal-2022-multimodal,shen2021text}. 
However, video-based dialogue generation remains an underexplored area and presents considerably greater challenges. 
It requires models with advanced video understanding and reasoning capabilities to achieve a fine-grained comprehension of the content, enabling the generation of dialogues that accurately reflect the interactions within the video scene and events.
Generating dialogue based on video content has broad practical applications, such as video re-creation and film dubbing, which are in high demand by video creators on popular platforms like TikTok and YouTube Shorts.

In this paper, we present a new task called \textbf{Theme-aware Video Dialogue Crafting (TVDC)}, which aims to generate a new dialogue that aligns with video content and follows user-specified themes or conditions, as shown in Figure~\ref{fig:intro}.
Conventional dialogue crafting for videos is a demanding and labor-intensive endeavor requiring substantial human involvement, including multiple video viewings and extensive revisions.
In contrast, automatically generating dialogue significantly alleviates the burden on video creators, allowing them to quickly obtain high-quality new dialogues that match the theme.
For instance, we can craft a new dialogue related to voting for a video originally themed around asking for directions.
There are two major challenges in TVDC: 
(1) \textbf{Theme Alignment.} The dialogue between characters should revolve around the given theme.
(2) \textbf{Visual Consistency.} The content of the dialogue associated with the current role needs to be consistent with the visual scenes, such as facial expressions and body movements portrayed in the video.

However, it is a challenge to leverage existing methods to create new dialogues that align with both the theme and the video content.
CHAMPAGNE~\cite{han2023champagne} proposed a model for predicting the next dialogue sentence in a video, but it generates only a single response based on historical dialogue and cannot adapt to new themes. 
While multimodal large language models (MLLMs) like video-chatGPT~\cite{maaz-etal-2024-video} and PLLaVA~\cite{xu2024pllava} are effective at understanding general video content and can interact around user-specified themes, they struggle to generate new dialogues that align with both the visual content and theme. 
Moreover, they cannot accurately model the relationships and dialogue order among the characters in the video.
In addition, existing methods generate all dialogue at once, which leads to a lack of immersion and fails to precisely align with the fine-grained expressions of characters at different periods in the video. 
To achieve better dialogue quality, we believe each character should have autonomy and the ability to think independently.

Therefore, we propose a multimodal agent method based on LLMs for \textbf{T}heme-aware \textbf{V}ideo \textbf{Dialogue} generation, focusing on maintaining \textbf{T}heme alignment and \textbf{V}isual consistency, called \textbf{TV-Dialogue}.
It enables immersive interaction among roles while dynamically rectifying the generated dialogue, thereby facilitating the accurate and efficient creation of dialogue that satisfies the specified themes. 
Additionally, our method can handle videos of \textbf{ANY} length and \textbf{ANY} open-world themes in zero-shot without training.
Specifically, TV-Dialogue begins by creating a new plot based on the user-specified theme and the original video, assigning new roles to each character as sub-agents.
During each character's dialogue period in the video, the corresponding sub-agent utilizes the visual-language model (VLM)~\cite{liu2023llava} to perceive its own visual behaviors and emotional changes. 
It then combines contextual information and messages from other sub-agents to generate new dialogue that aligns with the current context.
Next, TV-Dialogue employs a dialogue self-correction mechanism to evaluate whether the generated dialogue meets the criteria. It provides revision suggestions and allows the sub-agent to regenerate the dialogue. Finally, the corresponding sub-agent updates its own state and sends the generated dialogue sentence to the other sub-agents.
In addition, we propose a multi-granularity evaluation benchmark that assesses the quality of generated dialogues by jointly providing evaluation scores and comments, thereby enhancing the accuracy and reliability of the GPT-based evaluation mechanism.
We conducted extensive experiments using our self-collected Multi-Theme Video Dialogue (MVD) dataset, demonstrating the superior effectiveness of TV-Dialogue compared to commercial GPT models and multimodal large language models.
Meanwhile, we demonstrated the advantages of the new dialogues generated by TV-Dialogue for downstream multimodal tasks, such as video-text retrieval.
Experimental results show that pre-training with our generated new dialogues can effectively improve performance on the video-text retrieval task, exceeding baseline methods by more than 6\% in recall at rank 5 (R@5).

Our contributions are summarized as follows:
\begin{itemize}
\item We introduce a new task, Theme-aware Video Dialogue Crafting (TVDC), 
to generate dialogues aligned with the video content and user-specified themes.
\item We introduce a multimodal agent framework TV-Dialogue, which generates new dialogues on ANY theme for videos of ANY length. 
It achieves real-time immersive interaction, allowing each character in the video to perceive the current environmental information from a first-person perspective and engage in dialogue with other characters.
\item We establish a multi-granularity evaluation benchmark for TVDC, evaluating the quality of generated dialogues by jointly providing evaluation scores and assessments.
The benchmark evaluates the quality of generated dialogue from both text and video dimensions, ensuring high accuracy, interpretability, and reliability.
\end{itemize}

\section{Related Work}

\subsection{Multimodal Dialogue Generation}
Existing multimodal dialogue generation~\cite{sun-etal-2022-multimodal,yang2021open} can be categorized into two types based on the modalities employed: text and vision.
Text-based dialogue generation~\cite{chen2017survey,ni2023recent} is one of the most classic tasks and relies solely on text information for dialogue.
For example, \cite{vinyals2015neural} introduced a Seq2seq framework, which predicts the next sentence based on the previous sentence in a conversation.
Building on this, a substantial amount of work~\cite{huang2018automatic,song2019generating, li2022knowledge} has focused on forcing dialogue generation to express emotions. 
The rapid development of LLMs has provided a new perspective on dialogue generation. AutoGen~\cite{wu2023autogen} incorporates the paradigms of conversable agents and conversation programming. DiagGPT~\cite{cao2023diaggpt} constructs a multi-agent and collaborative system, extending more task-oriented dialogue scenarios.
Vision-based dialogue generation~\cite{liu2022survey,liao2018knowledge,sundar-heck-2022-multimodal,chen2020multimodal} extends text-based dialogue by incorporating images~\cite{zheng-etal-2022-mmchat,shuster-etal-2020-image} or videos~\cite{zhao-etal-2022-m3ed} into the conversation.
MMDialog~\cite{feng-etal-2023-mmdialog} introduced a multi-turn dataset containing images and proposed multimodal response generation and retrieval baselines. 
TikTalk~\cite{lin2023tiktalk} introduced a video-based multimodal chitchat task to facilitate dialogue with multimodal context.
Unlike the above approaches that use images or video as the focus of discussion, Champagne~\cite{han2023champagne} takes the video title, image frames, and history dialogue as input to predict the next sentence of dialogue in a video. 
However, Champagne can only predict a single sentence based on prior dialogue, and it neither generates entirely new dialogue for all characters nor creates dialogue according to any given theme.

\subsection{Dialogue Evaluation}
Accurately and comprehensively evaluating the quality of generated dialogue remains an unresolved challenge~\cite{li2021evaluate}. 
We categorize existing evaluation methods into automated evaluation~\cite{hastie2012metrics,tao2018ruber}, human evaluation~\cite{cummins2018neural,venkatesh2018evaluating}, and LLM-based evaluation~\cite{ou-etal-2024-dialogbench}.
For automated evaluation, a straightforward approach is to borrow evaluation metrics from other NLP tasks, such as BLEU~\cite{papinesi2002bleu}, ROUGE-L~\cite{lin2004rouge}, and METEOR~\cite{banerjee2005meteor}, which are widely used in dialogue prediction with a standard response. 
Another approach is to calculate the similarity between generated dialogue and reference dialogue, using metrics such as BERTScore~\cite{zhang2019bertscore}, BLEURT~\cite{sellam-etal-2020-bleurt}, and RoBERTa-eva~\cite{zhao-etal-2020-designing}. 
However, these methods require ground truth as a reference and  assess from only a single perspective.
Although human evaluation is a reliable method~\cite{see-etal-2019-makes}, it is inefficient, highly subjective, and prone to variation between evaluators~\cite{howcroft2020twenty}.
The rapid development of LLMs has made it possible to simulate human-like evaluation~\cite{chen2024llm, zhang2024comprehensive}.
DIALEVALML~\cite{mendonca-etal-2023-simple} proposes a reference-free dialogue evaluation framework that leverages strong pretrained LLM.
LLM-EVAL~\cite{lin-chen-2023-llm}  streamlines the evaluation process by using a single prompt and a
unified evaluation schema.
However, these frameworks cannot be directly applied to the TVDC task, as they lack the ability to incorporate visual information and assess the relationship between dialogue and visual content.

\section{Methodology}

\begin{figure*}[t]
    \centering    \includegraphics[width=\textwidth]{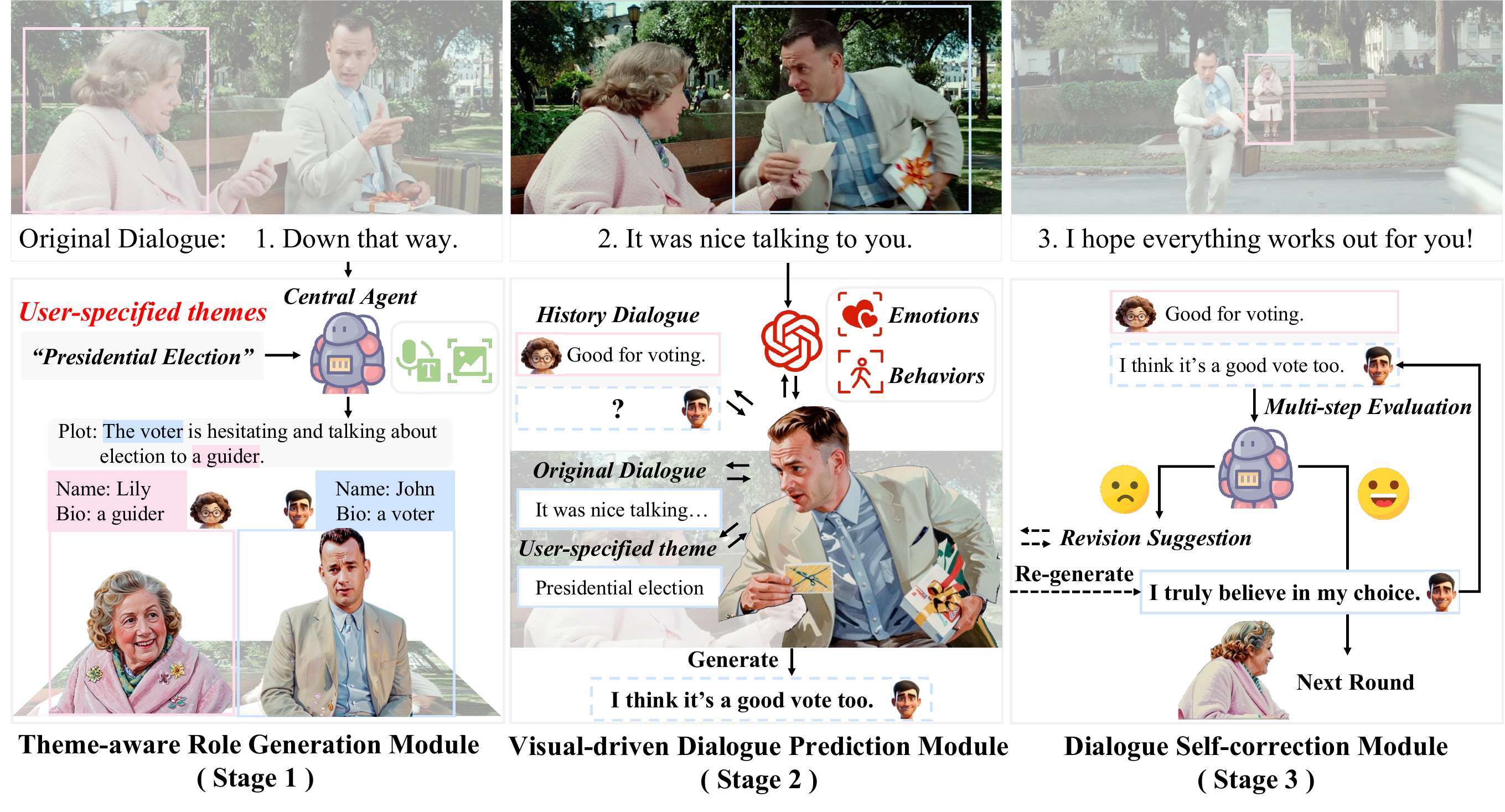}
    \vspace{-0.3in}
    \caption{Overview of TV-Dialogue. The TV-Dialogue initially assigns a relevant role to each sub-agent based on the given theme and video, enabling immersive interaction among the sub-agents in the dialogue process (Stage 1). Sub-agents maintain visual consistency by perceiving video content, querying historical memory, and receiving messages from other agents, thereby generating high-quality dialogues (Stage 2). The generated dialogues undergo self-correction for further improvement (Stage 3).}
    \vspace{-0.2in}
    \label{fig:overview}
\end{figure*}

\subsection{Overview}
The theme-aware video dialogue crafting (TVDC) task aims to generate a new dialogue that aligns with the given video and user-specified theme.
We present TV-Dialogue to address the TVDC task, focusing on theme alignment and visual consistency.
TV-Dialogue simulates human cognitive and communication processes by engaging in dialogue from a first-person perspective centered around the specified theme, thereby accurately understanding the video content and generating a new dialogue that aligns with the given themes.
TV-Dialogue consists of a central-agent $A_{0}$ as a core to manage the dialogue process, along with other sub-agents $A_{i}$ as dialogue participants, where $i = 1, \ldots, k$, and $k$ denotes the number of characters involved in the conversation within the video.
The central-agent initially comprehends the video $V$ and the given theme $C$, subsequently creating a new plot related to the theme and assigning corresponding new roles to each sub-agent.
Consequently, each sub-agent engages in immersive interaction with the other roles in the video from a first-person perspective (Sec.~\ref{method:memory_construct}).
These sub-agents need to perceive their own emotions and behaviors in real time, and predict the current dialogue based on their characteristics and historical dialogue information (Sec.~\ref{method:agent_communicating}).
The central-agent will assess whether the generated dialogue aligns with the context, offering revision suggestions and prompting the sub-agent to regenerate it if it does not meet the criteria (Sec.~\ref{method:self_reflect}).
TV-Dialogue can leverage any large language model as the core to execute the above process (Figure~\ref{fig:overview}).

\subsection{Theme-aware Role Generation Module}\label{method:memory_construct}
To generate new dialogues that align with user-specified themes, TV-Dialogue aims to simulate human conversational behavior.
Based on the theme, TV-Dialogue proposes creating a new plot for the video and assigning each character a new role.
Each character, using its new role, engages in immersive interaction from a first-person perspective, thereby generating high-quality and theme-aligned dialogues.
To achieve this goal, TV-Dialogue obtains the first frame of the video and inputs it into Vision-Language Models (VLM) for visual understanding, thereby obtaining a rough description of the entire video.
Meanwhile, It also employs the Automatic Speech Recognition (ASR) algorithm~\cite{radford2023robust} to recognize the content of original dialogues from the video.
Next, the central-agent utilizes the above information to generate a new plot, creates new roles, and assign them to each sub-agent $A_{i}$, who acts as a real character in the video.
Each sub-agent $A_{i}$ has a state, which includes its inherent information $\mathrm{role}_{i}$ (role name, description, etc.) and $memory_{i}^t$. Here, $memory_{i}^{t}$ contains the historically generated dialogue and the original dialogue content, which is dynamically updated as the conversation progresses.
We formulate the state $s_{i}^{t}$ of the $i$-th sub-agent in $t$-th round of conversation as:
\begin{equation}
    s_{i}^{t} = [\mathrm{role}_{i}; memory_{i}^t],
\end{equation}
where in the initial state, $t$ equals to 0 represents the initial state, and $memory^0$ is empty. 
At this point, each character in the video engages in immersive interaction using their roles based on the theme, resulting in dialogue content that is highly aligned with the user-specified theme.

\subsection{Visual-driven Dialogue Prediction Module}
\label{method:agent_communicating}
During the conversation period, the fine-grained visual expressions among the characters in the video guide the dialogue generation process. 
The dialogue generated by a character needs to be consistent with the corresponding visual representation in the video during that period. For example, the dialogue ``I am very happy to publish a paper'' should correspond to a student in the video who is laughing.
To generate dialogues that align with the visual expression in the video, each sub-agent needs to maintain visual consistency in each round of conversation, i.e., the generated content should be consistent with the emotions and behaviors of the current role.

Specifically, the sub-agent $A_{i}$ first perceives the situation of the character in the $t$-th round of conversation, capturing visual cues such as behaviors $a_{i}^{t}$ and emotions $e_{i}^{t}$ to provide strong prior for next-step. 
It employs MLLMs to conveniently obtain multimodal information at different granularities by adjusting  input prompt. This approach contrasts significantly with previous works such as~\cite{fan2024videoagent, wang2024videoagent}, which requires numerous specialized models, resulting in the framework being inflexible and greatly reducing generalization.

Subsequently, the sub-agent $A_{i}$ reads the recorded information in $memory_{i}^{t-1}$, and receives the generated dialogue sentence $d_{t-1}$ from the last speaking agent.
Finally, based on the state $s_{i}^{t-1}$ of $A_{i}$, as well as current behaviors and emotional information, we predict the generated dialogue sentence $d_{t}$ of the current $t$-th round of conversation as:
\begin{equation}
    d_{t} = A_{i}(s_{i}^{t-1}, a_{i}^{t}, e_{i}^{t}, d_{t-1}).
\end{equation}
It ensures consistency between the generated dialogue and the visual expression of the corresponding character in the video during that period.
Subsequently, the sub-agent updates the generated dialogue sentence $d_{t}$ into $memory_{i}^{t}$ and obtains a new state $s_{i}^{t}$. 
To further enhance the visual consistency of dialogue generated by sub-agent $A_{i}$, we adopt Plan-and-Solve prompting mechanism~\cite{wang-etal-2023-plan} to complete the above process.

\subsection{Dialogue Self-correction Module}\label{method:self_reflect}

Although significant efforts have been made to maintain theme alignment and visual consistency, the dialogue generated by sub-agent $A_{i}$ might still contain inappropriate content. 
For example, the sentence may be inconsistent with previous dialogue or too long to fit within the video’s time constraints. 
In such cases, the central-agent $A_{0}$ will evaluate the quality of the generated dialogue and decide whether the sub-agent $A_{i}$ needs to regenerate it.
We introduce a more lenient assessment method that provides constructive and effective revision suggestions for the currently generated content, rather than simply judging confidence scores. 

The central-agent evaluates the generated content from a local to a global perspective to determine its suitability and offers revision suggestions. 
Specifically, the central-agent $A_{0}$ first assesses whether the current dialogue $d_{t}$ is thematically aligned and contextually appropriate.
Next, it evaluates the content and logical continuity between the current dialogue $d_{t}$ and the previous round $d_{t-1}$.
Finally, the central-agent $A_{0}$ checks the overall coherence of the current dialogue $d_{t}$ in relation to the historical dialogue $d_{1}, ..., d_{t-1}$.
After the assessment process, if the central-agent $A_{0}$ determines that the generated dialogue meets the criteria, it will be accepted as the response of the current sub-agent $A_{i}$. 
Otherwise, it will output revision suggestions $o_{t}$. 
These suggestions reflect the feedback of central-agent $A_{0}$ on the current dialogue $d_{t}$, highlighting issues and providing directions for improvement.
The sub-agent $A_{i}$ needs to consider the revision suggestions and re-generate the dialogue:
\begin{equation}
    d_{t}^{'} = A_{i}(s_{i}^{t-1}, a_{i}^{t}, e_{i}^{t}, d_{t-1}, o_{t}).
\end{equation}
We summarize TV-Dialogue as Algorithm~\ref{alg:algorithm}.

\begin{table}[!t]
\centering
\begin{minipage}[b]{0.48\linewidth}
\caption{Statistics information of the MVD dataset.}
\centering
 \resizebox{\textwidth}{!}{
\begin{tabular}{lc}
\hline
Statistic                    &       \\ \hline\hline
\# videos                    & 351   \\
avg. \# role numbers         & 2.19  \\
avg. \# dialogue turn        & 3.75  \\
avg. dialogue length (words)         & 35.13 \\
avg. video length (seconds)           & 16.49 \\ \hline
\end{tabular}
}
\vspace{-0.2in}
\label{tab:statics}
\end{minipage}
\hfill
\begin{minipage}[b]{0.48\linewidth}
\centering
\includegraphics[width=\linewidth]{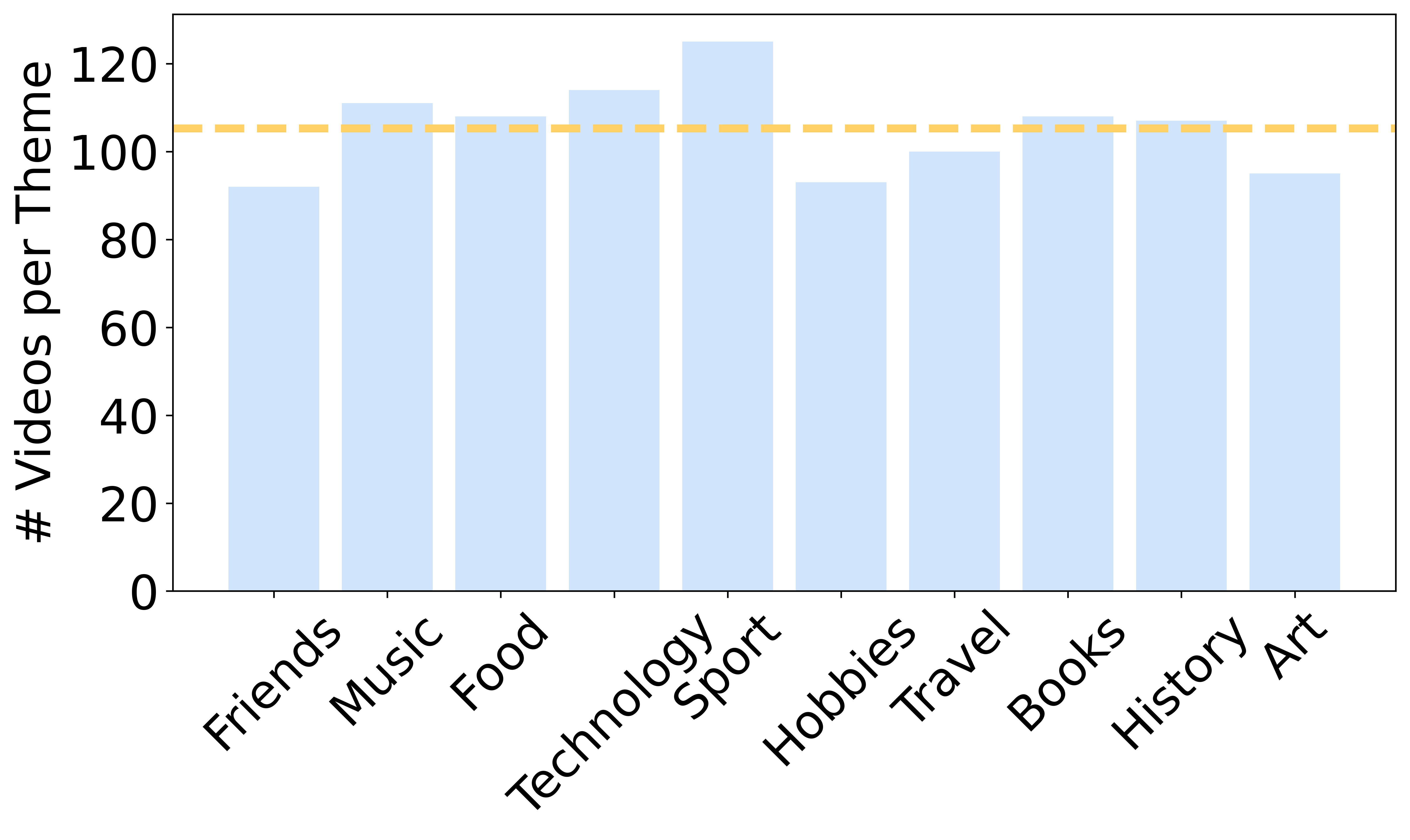}
\vspace{-0.4in}
\captionof{figure}{Videos per theme.}
\vspace{-0.2in}
\label{fig:themestatics}
\end{minipage}
\end{table}

\begin{algorithm}[t]
\small
\caption{TV-Dialogue}
\label{alg:algorithm}
\SetAlgoLined

\KwData{Video $V$, User-specified theme $C$, LLM $F_{l}$, VLM $F_{v}$, dialogue round $T$, max iteration $N$} 

\KwResult{Generated Dialogue $D$}

\tcp*[h]{\fontsize{8}{0}\selectfont \colorbox{YellowGreen}{\textnormal{\textbf{Stage1: Theme-aware Role Generation Module}}}}

Initialize $D$ = $\emptyset$ 

Slicing video into T segments: $V = [v_{1}, v_{2},...,v_{T}]$

$A_{0}$ $\leftarrow$ \text{Initialize the central-agent with }  $F_{l}$

plot, role~$\leftarrow$~\text{Generate new plot and roles with} $A_{0}$, $C$, $v_{1}$

$s_{1}^{t}, ..., s_{k}^{t}$ $\leftarrow$ \text{Obtain initial state with }$A_{0}$, plot, role

$A_{1}$,..., $A_{k}$ $\leftarrow$ \text{Initialize sub-agents with }$F_{l}$, $s_{1}^{t}$, $\dots$, $s_{k}^{t}$

\For{$t\leftarrow 1$ \KwTo $T$}{ 
\tcp*[h]{\fontsize{8}{0}\selectfont \colorbox{YellowGreen}
{\textnormal{\textbf{Stage2: Visual-driven Dialogue Prediction Module}}}}
    
    $a_{i}^{t}$,~$e_{i}^{t}$ $\leftarrow$ \text{Obtain behavior, emotion with }$A_{i}$, $F_{v}$, $v[t]$
    
    $d_{t-1}$ $\leftarrow$ \text{Query history dialogue from }$s_{i}^{t-1}$ 
    
    \For{$n\leftarrow 1$ \KwTo $N$}{ 
        
        $o_{t}$ $\leftarrow$ \text{Query suggestion from }$A_{0}$
        
        $d_{t}$  $\leftarrow$ \text{Generate dialogue with } $A_{i}$, $s_{i}^{t-1}$, $a_{i}^{t}$, $e_{i}^{t}$, $d_{t-1}$, $o_{t}$
        
\tcp*[h]{\fontsize{8}{0}\selectfont \colorbox{YellowGreen}{\textnormal{\textbf{Stage3: Dialogue Self-correction Module}}}}
        
        $o_{t}$ $\leftarrow$ \text{Evaluate $d_{t}$ and obtain suggestions with }$A_{0}$
        
        \If{$o_{t}$ \text{is empty}}{
        
            break\;
        }
    }
    
    $D \leftarrow D \cup d_{t}$
}

\textbf{return} $D$

\end{algorithm}

\begin{table*}[htb]
\vspace{-0.2in}
\small
\centering
\caption{The evaluation metrics with corresponding definitions.}
\begin{tabular}{cl|l}
\hline
\multicolumn{2}{c|}{Metric}                                                              & \multicolumn{1}{c}{Definition}                                                                 \\ \hline\hline
\multicolumn{1}{c|}{\multirow{4}{*}{Text}}  & (1) Theme Relevance ($\mathbb{TR}$)  & The relevance of dialogues to the given theme.                            \\
\multicolumn{1}{c|}{}                       & (2) Generation Quality ($\mathbb{GQ}$)    & The fluency, grammar, and colloquialism of dialogues.                     \\
\multicolumn{1}{c|}{}                       &  (3) Logical Coherence ($\mathbb{LC}$)    & The logical coherence and reasonableness of dialogues.                    \\
\multicolumn{1}{c|}{}                       &  (4) Content Diversity ($\mathbb{CD}$)    & The diversity between generated dialogue and original dialogue.           \\ \hline
\multicolumn{1}{c|}{\multirow{2}{*}{Video}} & (5) Video Compatibility ($\mathbb{VC}$)   & The compatibility between dialogues and characters' behaviors and emotions. \\
\multicolumn{1}{c|}{}                       & (6) Scenario Consistency ($\mathbb{SC}$)  & The consistency between dialogues and video scenarios.                    \\ \hline
\end{tabular}
\vspace{-0.1in}
\label{tab:metric}
\end{table*}

\begin{table*}[ht]
\vspace{-0.1in}
\centering
\caption{Comparison with state-of-the-art methods on the MVD dataset. “TV-Dialogue ($\cdot$)” refers to the TV-Dialogue method we proposed using different LLM models.}
\small
\begin{tabular}{c|l|cccc|cc|c}
\hline
\multirow{2}{*}{Methods} & \multicolumn{1}{|c|}{\multirow{2}{*}{Model}} & \multicolumn{4}{c|}{Text-oriented} & \multicolumn{2}{c|}{Video-oriented} & \multirow{2}{*}{Average} \\ \cline{3-8}
                         &                        & $\mathbb{TR}$      & $\mathbb{GQ}$      & $\mathbb{LC}$      & $\mathbb{CD}$      & $\mathbb{VC}$                & $\mathbb{SC}$                &                          \\ \hline\hline
\multirow{2}{*}{Text}    & GPT-3.5~\cite{ouyang2022training}                & 3.82    & 3.19   & 2.74   & 3.86   & 2.91             & 3.13             & 3.28                     \\
                         & GPT-4o~\cite{openai2024hello}                 & 3.82    & 3.26   & 2.83   & 3.86   & 2.93             & 3.12             & 3.30                     \\ \hline
Image                    & GPT-4V~\cite{achiam2023gpt}                 & 3.82        & 3.32       & 2.92       &  3.74      & 3.02                 &  3.16               &  3.33                        \\ \hline
Video & PLLaVA~\cite{xu2024pllava}                 & 3.39        & 3.50       & 3.04       & 2.93       & 2.88                 & 3.29             & 3.17                         \\ \hline
\multirow{5}{*}{Ours}   
                         & \multicolumn{1}{l|}{TV-Dialogue (GLM~\cite{glm2024chatglm})}            &  3.60       & 3.68       & 3.24       & 3.88       &  2.93                &   3.11               &  3.41                        \\
                         & \multicolumn{1}{l|}{TV-Dialogue (QWen-2.5~\cite{bai2023qwen})}                  & 3.90        & 3.71        & 3.43       & 4.13       & 3.08                 &  3.12                &  3.56          \\
                         & \multicolumn{1}{l|}{TV-Dialogue (LLama-3.1~\cite{touvron2023llama})}              & 3.71    & 3.54   & 3.11   & 3.80   & 3.08             & 3.08             & 3.39                     \\
                         & \multicolumn{1}{l|}{TV-Dialogue (GPT-3.5)}           & 4.03    & 3.87   & 3.73   & 4.42   & \textbf{3.19}             & 3.26             & 3.75                     \\
                         & \multicolumn{1}{l|}{TV-Dialogue (GPT-4o)}           & \textbf{4.17}    & \textbf{4.07}   & \textbf{3.94}   & \textbf{4.45}   & 3.11             & \textbf{3.31}             & \textbf{3.84}                     \\ \hline
\end{tabular}
\vspace{-0.2in}
\label{tab:result}
\end{table*}

\section{Experiment}
\subsection{The Multi-Theme Video Dialogue Dataset}
To better validate the capability to generate new dialogues for videos, particularly under multi-theme conditions, we propose a dedicated video dialogue dataset specifically for TVDC.
Specifically, we propose the Multi-Theme Video Dialogue dataset \textbf{(MVD)}, consisting of 351 dialogue-intensive video clips that span various application scenarios, including movie scenes, daily life interactions, and cartoon settings. 
The detailed statistics are shown Table~\ref{tab:statics}.
The MVD dataset focuses on conversations among multiple participants, and the selected video content is highly scalable, allowing for dialogues that are not limited to any specific theme.
These clips are manually extracted from YouTube, mainly focusing on character dialogue scenes with diverse dialogue content and rich visual events. 
The characters in the videos we selected have vivid facial expressions and diverse body movements, but they do not possess explicit identity and scene attributes.
Therefore, they have high flexibility and can be matched with a wide range of new themes. 
Additionally, we have carefully selected 10 conversation themes frequently encountered in daily life, such as ``Music'' and ``Friends'', ensuring that they are minimally influenced by the video context in which the dialogue takes place. 
We randomly assign three themes to each video for evaluation, and the number of videos corresponding to each theme is shown in Figure~\ref{fig:themestatics}.

\subsection{Evaluation Metric}\label{metric:agent_metric}
To achieve a comprehensive and reliable evaluation of the generated dialogues, we use both traditional metrics and qualitative assessments based on the proposed multi-granularity evaluation benchmark.

For traditional metrics, we employ BertScore, METEOR, and ROUGE-L to ensure objective, quantifiable assessments. 
Although traditional metrics can quantitatively assess sentence patterns and N-gram similarity between generated and reference dialogues, the TVDC task does not have a ground truth (i.e., reference dialogue) for new dialogues generated based on user-defined themes and video content.
Therefore, we only apply traditional metrics to the Last K Sentence Prediction study.

In the absence of ground truth dialogues, we designed a series of qualitative evaluation metrics across two dimensions: text-oriented (1-4) and video-oriented (5-6), to more accurately and reliably assess the quality of generated dialogues, particularly in terms of theme alignment and visual consistency.
As shown in Table~\ref{tab:metric}, we define all evaluation metrics as soft constraints, scored on a 1 to 5 point scale, with 5 being the best.
Each metric includes a corresponding definition with detailed evaluation criteria. 
Additionally, we develop an evaluation pipeline using the GPT-4o-mini model, which assigns evaluation scores along with comments, thereby enhancing the accuracy and reliability of evaluation. 
We set the decoding temperature of the evaluation model to 0 to further increase determinism, resulting in a standard deviation of evaluation results of less than 0.01, thereby ensuring the credibility of the findings.
It should be noted that changing the evaluation model will not affect the evaluation results.

\subsection{Implementation Details}\label{sec:implementation}

We use Whisper (medium.en)~\cite{radford2023robust} for speech recognition.
PLLaVA (7B)~\cite{xu2024pllava} serves as the VLM and the only external visual tool, capturing information at various levels of granularity by adjusting prompts.
It should be noted that it can be substituted with other visual language models.
Our approach supports any large language model as the core. 

\begin{table*}[ht]
\vspace{-0.1in}
\caption{Ablations on different modules and external information used in TV-Dialogue on the MVD dataset. 
``Role'' represents the theme-aware role generation module.
``Visual'' means the visual-driven dialogue prediction module, which is also responsible for incorporating external information (i.e., emotion and action).
``Correction'' denotes the dialogue self-correction module.
}
\small
\centering
\begin{tabular}{c|ccc|cc|cccc|cc|c}
\hline
\multirow{2}{*}{Group} & \multicolumn{3}{c|}{Module}           & \multicolumn{2}{c|}{Information} & \multicolumn{4}{c|}{Text-oriented} & \multicolumn{2}{c|}{Video-oriented} & \multirow{2}{*}{Average} \\ \cline{2-12}
   & Role  & Visual & Correction & Emotion         & Behavior         & $\mathbb{TR}$      & $\mathbb{GQ}$      & $\mathbb{LC}$   & \multicolumn{1}{c|}{$\mathbb{CD}$} & $\mathbb{VC}$               & $\mathbb{SC}$               &                          \\ \hline\hline
\multirow{3}{*}{$\mathbb{A}$}  &  \checkmark    &        &            &                 &                & 3.92    & 3.91   & 3.68   & 3.98   & 3.08             & 3.34             & 3.65                     \\
  & \checkmark    & \checkmark      &            & \checkmark               & \checkmark              & 3.98    & 3.86   & 3.69   & 4.37   & 3.14             & 3.24             & 3.71                     \\
  &  \checkmark    & \checkmark      & \checkmark          & \checkmark               & \checkmark              & 4.03    & 3.87   & 3.73   & 4.42   & 3.19             & 3.26             & 3.75                     \\ \hline
\multirow{2}{*}{$\mathbb{B}$}  &  \checkmark    & \checkmark      & \checkmark          & \checkmark               &                & 3.60     & 3.83   & 3.65   & 3.77   & 3.38             & 3.28             & 3.58                     \\
   & \checkmark    & \checkmark      & \checkmark          &                 & \checkmark              & 3.83    & 3.86   & 3.73   & 3.81   & 3.09             & 3.31             & 3.61                     \\ \hline
\end{tabular}
\vspace{-0.2in}
\label{tab:ablation}
\end{table*}

\subsection{Comparison Methods}
Considering that there are currently no methods for theme-aware video dialogue generation, we selected a series of mainstream commercial and open-source large language models and designed baseline methods for comparison. 
These models represent the state-of-the-art in the fields of text, image, and video comprehension, respectively.
Specifically, we chose the commercial models GPT-3.5~\cite{ouyang2022training}, GPT-4o, and GPT-4V~\cite{achiam2023gpt}, as well as the open-source models PLLaVA (7B)~\cite{xu2024pllava}, LLaMA-3.1 (8B)~\cite{touvron2023llama}, QWen-2.5 (7B)~\cite{bai2023qwen}, and GLM (9B)~\cite{glm2024chatglm}).

The experiments consist of three types. 
Following Section~\ref{method:memory_construct}, we first generate a new plot and roles based on the theme, which will be used in all subsequent methods.
For text-based methods, we treat video dialogue generation as a purely linguistic task. 
We utilize the VLM to convert information about the behaviors and emotions of characters during different conversation periods into text, allowing the LLM to generate new dialogue based on the given theme all at once.
For image-based methods, we further integrate key image frames as an additional modality, building on the text-based approach.
For video-based methods, since MLLMs can directly process video information, we input the video along with the generated plot and roles to directly generate new dialogue.

\subsection{Comparison with State-of-the-arts}\label{sec:comparison}

Table~\ref{tab:result} shows the performances of various models on the MVD dataset.
The method based on TV-Dialogue consistently outperformed other approaches across all metrics, achieving better results than both end-to-end MLLMs and their corresponding LLM counterparts. 
Moreover, it struck the best balance between text and video dimensions. 
Even in cases where there were significant parameter differences between LLMs, such as LLama-3.1 (8B) compared to GPT-4o, the TV-Dialogue version of LLama-3.1 still outperformed the text-based GPT-4o.

In contrast, the dialogue generated by text-based methods clearly exhibits lower logical coherence and content diversity. 
They tend to rigidly align with the original dialogue of the video based on themes, making it challenging to generate new dialogues that reflect the expressions of specific characters.
This is because they generate all new dialogue at once, rather than modeling the individual characters separately, as TV-Dialogue does.
As the video length increases, the issue of losing fine-grained information becomes even more pronounced.
In the video-based method, although PLLaVA supports process video directly, it clearly lacks the generative capabilities required for TVDC.
In other words, PLLaVA seems to focus more on captioning and generating dialogues casually and freely based on video content, significantly deviating from the given theme.

\subsection{Ablation Studies}
In this section, we conduct ablation experiments to verify the effectiveness of each component and thoroughly analyze the efficacy of our proposed method.

\textbf{Effectiveness of proposed module.} Table~\ref{tab:ablation} shows the performances of various variants of TV-Dialogue, each equipped with GPT-3.5 as the core model.
After adding the visual-driven dialogue prediction module, TV-Dialogue gained the ability to perceive characters' emotions and behaviors, resulting in a corresponding improvement in video compatibility, with the average score further increasing to 3.71.
As the available external information increased, content diversity further improved by 0.39, indicating that the generated dialogue exhibited greater variety.
Additionally, owing to the dialogue self-correction mechanism, TV-Dialogue effectively improves the quality of generated dialogue. This enhancement is particularly notable in text-oriented metrics, resulting in substantial advancements across all metrics.
In addition, we explored the impact of different information sources (emotion and behavior) on the quality of dialogue. 
As shown in Table~\ref{tab:ablation}, introducing either emotion or behavior information alone effectively enhances the consistency between the dialogue and visual content. 
When both are simultaneously incorporated, the overall quality of the dialogue is further improved.

\begin{table}[t]
\centering
\caption{
Comparison of the sentence prediction performance.
``Last-k'' denotes predicting the last k sentences of the original dialogue.
``TV ($\cdot$)'' refers to the TV-Dialogue we proposed.
}
 \resizebox{0.48\textwidth}{!}{
\begin{tabular}{c|c|c|cc}
\hline
\multirow{2}{*}{Pred} & \multirow{2}{*}{Model} & Semantic-Level (\%) & \multicolumn{2}{c}{Word-Level (\%)} \\ \cline{3-5} 
                                   &                        & BertScore      & METEOR        & ROUGE-L        \\ \hline\hline
\multirow{4}{*}{Last-1}         & GPT-3.5          & 85.46           & 7.95          & 4.80                    \\
                                 & TV (GPT-3.5)               &   85.79          &  9.68          & 5.39          \\
                                   & GPT-4o            & 85.89           & 10.01          & 5.65                   \\
                                   & TV (GPT-4o)                 & 85.78            &  10.92          &  5.13       \\ \hline
\multirow{4}{*}{Last-2}         & GPT-3.5           & 85.42           & 7.23          & 3.91                   \\
                                & TV (GPT-3.5)                 &   85.60          &  8.51          &  4.96       \\
                                    & GPT-4o          & 85.66           & 8.86          & 3.99              \\
                                   & TV (GPT-4o)                 &  85.64           &  9.49          & 4.18       \\ \hline
\end{tabular}
}
\vspace{-0.1in}
\label{tab:lastk}
\end{table}

\begin{figure}[tp]
\centering    
\includegraphics[width=0.48\textwidth]{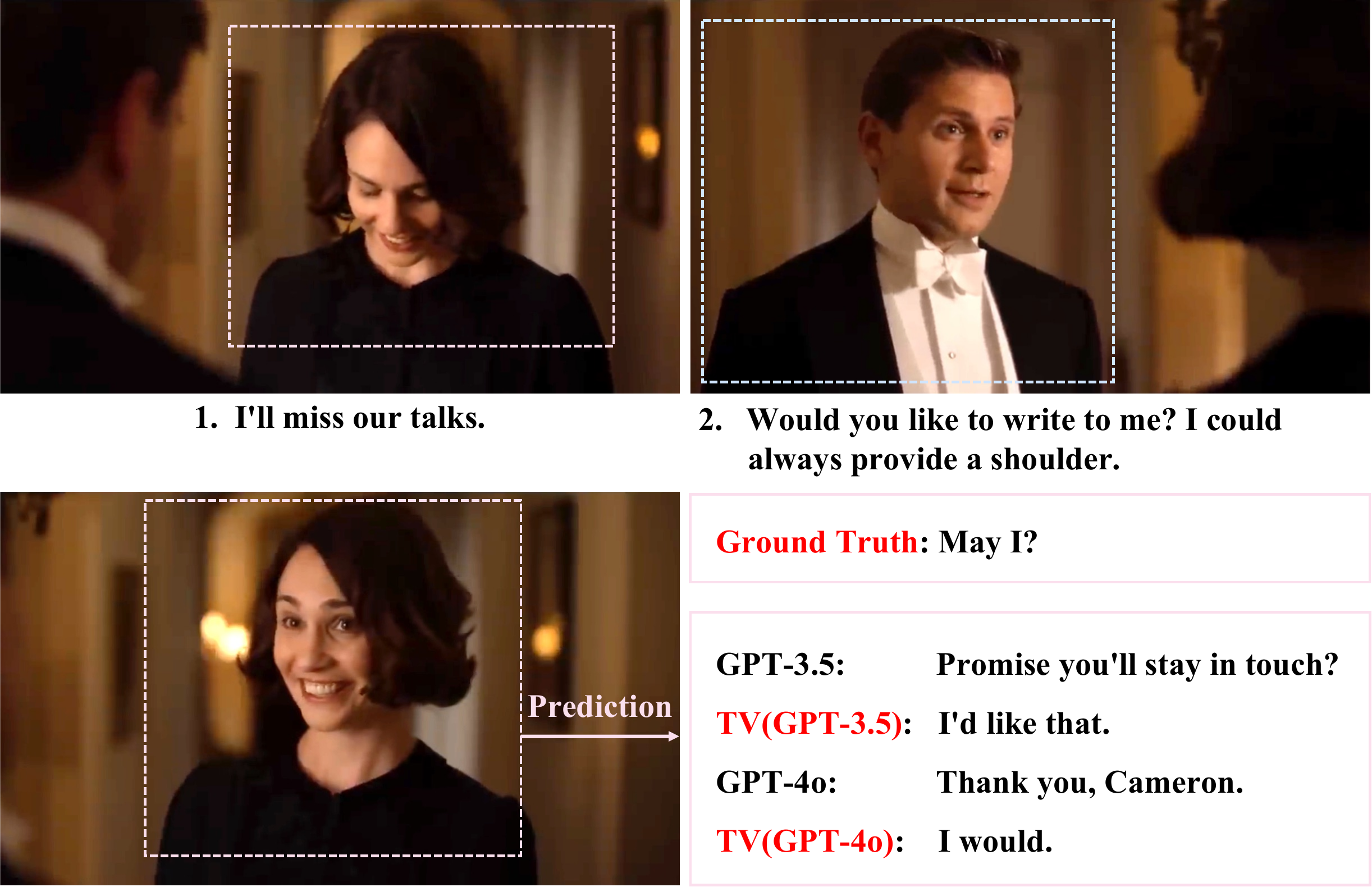}
\caption{
Comparison of dialogues generated by different methods in the last-1 sentence prediction. 
The top-left corner represents the first frame of the dialogue in the video.
Although the values of traditional metrics are very low, the generated dialogues are consistent with the video content and theme.
}    
\vspace{-0.1in}
\label{fig:lask_k}
\end{figure}

\textbf{Last K Sentence Prediction Study.} To validate the accuracy of the dialogue generated by TV-Dialogue, we task it with predicting the final K sentences of original dialogue based on the video content and the partial original dialogue. 
Since the original dialogue can be used as ground truth, we employ traditional metrics to assess the similarity between the predictions of TV-Dialogue and the original dialogue from two perspectives: semantic level (BertScore) and word level (METEOR and ROUGE-L).
First, we generate the corresponding plot and roles based on video and original dialogue. 
We then input the video and original dialogue before the last K sentences into TV-Dialogue for prediction. 

As shown in Table~\ref{tab:lastk}, TV-Dialogue and text-based methods achieve similar performance at both the semantic and word levels, with TV-Dialogue slightly outperforming the text-based methods. 
Both methods achieve relatively high semantic similarity, whereas word-level similarity is very low, with most values below 10\%.
This is because video dialogue generation may yield multiple suitable sentences for current conversation, making deterministic evaluation metrics unsuitable for the TVDC task. 
To further demonstrate the inadequacy of traditional metrics, we compare the differences between the generated dialogues and the ground truth sentences in Figure~\ref{fig:lask_k}.
It is evident that TV-Dialogue generates new dialogues with better theme alignment and visual consistency.
Therefore, compared to traditional metrics, the qualitative evaluation metrics proposed in Section~\ref{metric:agent_metric} are more suitable for evaluating the theme-aware video dialogue crafting task.

\textbf{Domain Transfer Study.} We aim to demonstrate the impact of high-quality new dialogues generated by TV-Dialogue on other downstream multimodal tasks.
We reference the classic video-text retrieval task, utilizing the original dialogues from the videos to retrieve the corresponding videos. 
We divide the 351 original videos  with dialogues on MVD into a training set and a test set, with 301 for training and 50 for testing. 
We train the classic Clip4clip~\cite{luo2022clip4clip} model on the training set following the default settings, and we use the recall at rank (R@K), Median Rank, and Mean Rank as evaluation metrics.
Additionally, we use the new dialogues generated by TV-Dialogue for the 301 original videos as supplementary data, jointly training with the original dialogues. 
As shown in Table~\ref{tab:retrieval}, after incorporating the dialogues generated by TV-Dialogue, most metrics achieve significant improvements, particularly with R@5 increasing by over 6\%. 
This indicates a strong correlation between the dialogue generated by TV-Dialogue and the video content.
It is important to note that TV-Dialogue can generate new dialogues for any video at no cost, making it possible to create large-scale pre-training datasets for downstream video-related or dialogue-related tasks.

\begin{table}[!t]
\vspace{-0.1in}
\caption{Comparison of video-text retrieval performance between the original dialogue (``Original'') and the new dialogue, denoted as ``New'', generated by TV-Dialogue (GPT-4o).}
 \resizebox{0.48\textwidth}{!}{
\begin{tabular}{c|ccccc}
\hline
\multicolumn{1}{c|}{Training Data} & R@1↑ & R@5↑ & R@10↑ & Median R↓ & Mean R↓ \\ \hline\hline
Original              & 16.0     & 32.0     & 46.0      & 13.5      & 15.6    \\
New                    & 12.0     & 28.0     & 46.0      & 12.0      & 16.1    \\
Original+New          & 16.0     & 38.0     & 48.0      & 13.0      & 14.3    \\ \hline
\end{tabular}
}
\vspace{-0.3in}
\label{tab:retrieval}
\end{table}

\begin{figure}[t]
\centering    
\includegraphics[width=0.24\textwidth]{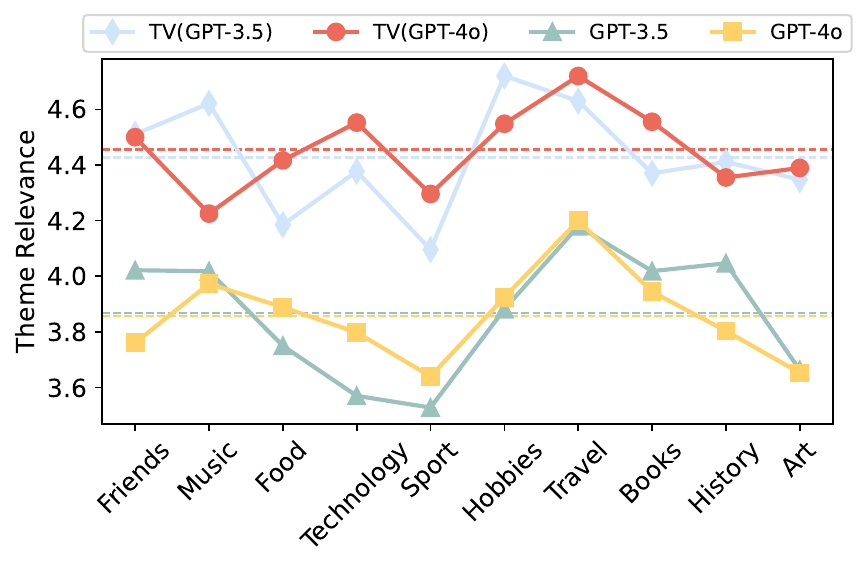}
\hspace{-0.1in}
\includegraphics[width=0.24\textwidth]{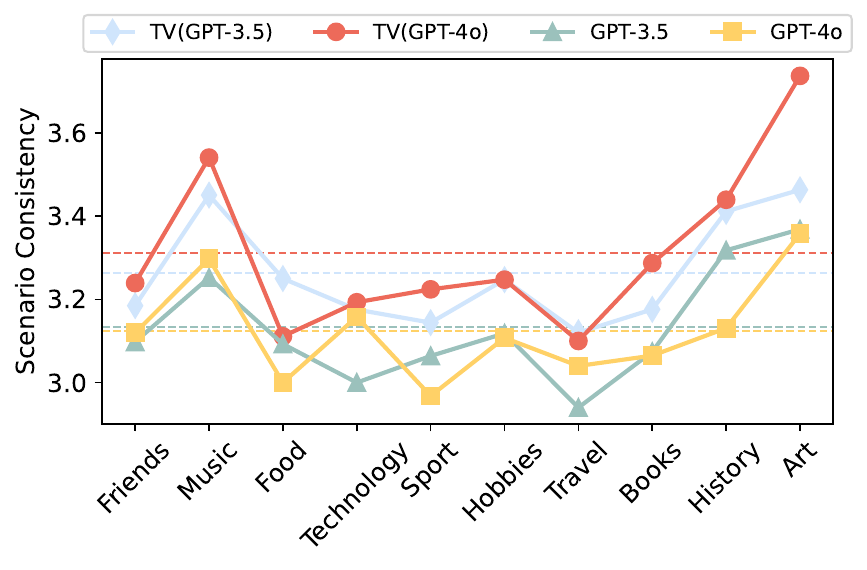}
\vspace{-0.2in}
\caption{
Comparison of different themes in terms of Theme Relevance ($\mathbb{TR}$) and Scenario Consistency ($\mathbb{SC}$).
}
\vspace{-0.3in}
\label{fig:theme}
\end{figure}

\textbf{Analysis on Different Themes.}
To investigate the impact of different themes on video dialogue generation, we compared the performance of various methods across different themes, specifically focusing on theme Relevance for theme alignment and scenario consistency for visual consistency.
As shown in Figure~\ref{fig:theme}, the overall trends of different methods on the same theme are similar, indicating that the choice of theme indeed affects the quality of the generated dialogues. 
For themes that significantly contradict the video content, such as discussing casual sports-related topics in a serious meeting scenario, it is evidently challenging to generate high-quality new dialogues that align with the theme.
However, TV-Dialogue consistently outperformed text-based methods across all themes, and the quality improved with the enhanced capabilities of LLM models. 
Additionally, the quality of dialogues produced by TV-Dialogue is less affected by theme variations, particularly in terms of theme Relevance, where the variance of TV-Dialogue (GPT-4o) across themes is only 0.019.

\begin{table}[!t]
\vspace{-0.1in}
\small
\caption{Pearson correlation coefficients between the proposed metrics and human ratings indicate a positive correlation trend.}
\centering
\begin{tabular}{c|ccc}
\hline
Metric              & Random  & $\mathbb{TR}/\mathcal{TR}$  & $\mathbb{VC}/\mathcal{VC}$ \\ \hline\hline
\textit{Pearson's r}    & 0.02     & 0.47    & 0.53   \\\hline
\end{tabular}
\vspace{-0.1in}
\label{tab:human}
\end{table}

\subsection{User Study}
To comprehensively and reliably evaluate the generated dialogue, we perform human studies on the results of TV-Dialogue and GPT-4o.
We collected 400 human feedback responses from 20 participants with no prior experience.
The participants were shown the generated dialogue by TV-Dialogue (GPT-4o) and GPT-4o with video, and were asked to choose the best one without knowing which specific one it was.
As a result, out of 400 selections, the dialogue generated by TV-Dialogue was chosen 290 times, accounting for 72.5\% of selections, overwhelmingly surpassing the GPT-4o. 
This suggests that the dialogue produced by our method aligns more closely with the video than that of GPT-4o.
In addition, we invited three participants from diverse backgrounds to rate the relevance of the generated dialogues to the theme ($\mathcal{TR}$) and their compatibility with the video ($\mathcal{VC}$) on a scale of 1 to 5 for all 50 videos.
Table~\ref{tab:human} shows the correlation between ratings provided by GPT evaluators ($\mathbb{TR}$ and $\mathbb{VC}$) and human evaluators ($\mathcal{TR}$ and $\mathcal{VC}$), indicating a positive correlation trend, further confirming the accuracy and reliability of our proposed multi-granularity evaluation benchmark.

\section{Conclusion}
In this paper, we introduce a novel task named theme-aware video dialogue crafting (TVDC) and propose the TV-Dialogue to create new dialogues that align with the given theme and video. 
Our key insight is to achieve immersive interaction, i.e., enabling different roles in the video to engage from a first-person perspective and foster conversations around the given theme.
Furthermore, we also established a multi-granularity evaluation benchmark with high accuracy, interpretability, and reliability.

\section*{Impact Statement}
This paper presents research aimed at advancing practical applications, such as video re-creation and film dubbing, which are currently in high demand by video platforms like YouTube Shorts. 
The work we propose not only holds significant economic value but also substantially reduces labor costs, making it a highly practical and impactful contribution to the society.

Additionally, we have observed that the samples generated by TV-Dialogue may, in certain cases, lead to misunderstandings of the original video content when specific themes or contexts are introduced. 
However, unlike deepfake, which can generate highly sophisticated fake images that are challenging for humans to replicate, it is still possible to manually alter the original video to create new voiceovers.
Thus, our efforts will not raise new ethical concerns and are dedicated to advancing technological contributions and promoting positive applications within the field.




\nocite{langley00}

\bibliography{example_paper}
\bibliographystyle{icml2025}

\newpage
\newpage
\appendix

\end{document}